\documentclass[runningheads]{llncs}
\usepackage{graphicx}
\usepackage{subcaption}
\usepackage{hyperref}
\usepackage{xcolor}

\begin{document}
\title{Explanation-by-Example Based on Item Response Theory}
%
%
\author{Lucas F. F. Cardoso\inst{1,5}\and
José de S. Ribeiro \inst{1,2} \and
Vitor C. A. Santos \inst{1,4}\and
Raíssa L. Silva\inst{3}\and
Marcelle P. Mota\inst{1}\and
Ricardo B. C. Prudêncio\inst{4} \and
Ronnie C. O. Alves\inst{5}}
%

\authorrunning{Cardoso, L. et al.}

\institute{ICEN, Universidade Federal do Pará, Belém, Brazil\\
\email{lucas.cardoso@icen.ufpa.br, mpmota@ufpa.br}\and
IFPA, Instituto Federal do Pará, Belém, Brazil \\
\email{jose.ribeiro@ifpa.edu.br} \and
IRMB, Université Montpellier, France \\
\email{r.lorenna@gmail.com} \and
CIn, Universidade Federal de Pernambuco, Recife, Brazil\\
\email{rbcp@cin.ufpe.br} \and
ITV, Instituto Tecnológico Vale, Belém, Brazil\\
\email{\{vitor.cirilo.santos,ronnie.alves\}@itv.org}}

%
%
\maketitle              
\begin{abstract}
Intelligent systems that use Machine Learning classification algorithms are increasingly common in everyday society. However, many systems use black-box models that do not have characteristics that allow for self-explanation of their predictions. This situation leads researchers in the field and society to the following question: How can I trust the prediction of a model I cannot understand? In this sense, XAI emerges as a field of AI that aims to create techniques capable of explaining the decisions of the classifier to the end-user. As a result, several techniques have emerged, such as Explanation-by-Example, which has a few initiatives consolidated by the community currently working with XAI. This research explores the Item Response Theory (IRT) as a tool to explaining the models and measuring the level of reliability of the Explanation-by-Example approach. To this end, four datasets with different levels of complexity were used, and the Random Forest model was used as a hypothesis test. From the test set, 83.8\% of the errors are from instances in which the IRT points out the model as unreliable.

\end{abstract}
\section{Introduction}
The expansion and increasing use of Artificial Intelligence (AI) systems creates advances that enable these systems to learn and make decisions on their own \cite{gunning2019darpa}. Thus, AI becomes increasingly common in everyday society by providing for simple or complex decisions in people's lives to be taken via intelligent systems. Such decisions range from recommending movies based on the user's preferences to diagnosing a disease based on patient's exams \cite{review_xai_2021}.

The question ``Can the decision made by a black-box model be trusted for a context-sensitive problem?'' has been asked not only by the scientific community, but also by the society as a whole. For example, in 2018 the General Data Protection Regulation was implemented in the European Union. It is geared at securing anyone the right to an explanation as to why an intelligent system made a given decision \cite{regulation2018general}. In this sense, for a continuous advance in AI applications, the entire community is faced with the barrier of model explainability \cite{gohel2021explainable,gunning2019darpa}. To address this issue, a new field of study is growing rapidly: Explained Artificial Intelligence (XAI). Developed by AI and Human Computer Interaction (HCI) researchers, XAI is a user-centric field of study aimed at developing techniques to make the functioning of these systems and models more transparent
and consequently more reliable \cite{arrieta2020explainable}. Recent research shows that the trust calibration on the models' decision is very important, since exaggerated or measured confidence can lead to critical problems depending on the  context \cite{trust}.

The models that have high success rates to solve real-world problems are usually of the black-box type. In other words, they are not easily explained and, therefore, applying XAI techniques is required so that they can be explained and then interpreted by the end user\cite{gohel2021explainable,arrieta2020explainable}. The emergence of XAI techniques based on different methodologies is a real fact today, but there are still many gaps in literatute. For example, XAI methods based on Explanation-by-Example in a model-agnostic fashion \footnote{Model-Agnostic: it does not depend on the type of model to be explained \cite{molnar2020interpretable_review}.} are still underexplored by the scientific community \cite{interpretabilidade_review2018,guidotti2018survey_review,molnar2020interpretable_review}.
Techniques based on Explanation-by-Example use previously known ou model-generated data instances to explain them, thus providing for a good understanding of this model and decisions thereof. This is a technique that may be natural for human beings, since humans seek to explain certain decisions they themselves make based on previously known examples and experiences \cite{arrieta2020explainable}.

This research explores a new measure of XAI based on the working principles of Item Response Theory (IRT), which is commonly used in psychometric tests to assess the performance of individuals on a set of items (e.g., questions) with different levels of difficulty \cite{baker2001basics}. To this end, the IRT was adapted for Machine Learning (ML) evaluation, treating classifiers as individuals and test instances as items \cite{martinez2019item}. In previous works \cite{martinez2019item,cardoso2020decoding} IRT was used to evaluate ML models and datasets for classification problems. By applying IRT concepts, the authors were able to provide new information about the data and the performance of the models in order to grant more robustness to the preexisting evaluation techniques. In addition, the IRT's main feature is to explore the individual's performance on a specific item and then compute the information about the individual's ability and item complexity in order to explain why a respondent got an item right or wrong. Thus, it is understood that IRT can be used as a means to comprehend the relationship between the performance of a model and the data, thus helping in explaining models and understanding the model's predictions at a local level.


Given the intrinsic characteristics of the IRT, it is understood that it can be fitted within the universe of techniques based on Explanation-by-Example. At the same time, the IRT also has concepts that allow to explain and interpret the model in general and to shed light on details not yet explored by other XAI techniques. 
Based on this motivation, this research work proposes the use of IRT as a new Explanation-by-Example approach, in a model-agnostic way, aiming at greater reliability on the model's decisions by the end user. For the experiment, 4 datasets were selected with different levels of complexity indicated by \cite{ribeiro2021does} with the Random Forest algorithm acting as the target of the explanation. The objective of this research is to explore how the concepts from the IRT can help to open the black-box and indicate the confidence of the model's prediction. 

The remainder of this paper is divided into the following sections: Section 2 provides a contextualization about XAI and IRT; Section 3 explains how IRT is applied to ML and then to XAI; Section 4 provides the results and discussions of the proposal presented herein; Section 5 carries the conclusion of the herein research and final considerations related thereof.

\section{Background}

\subsection{Explainable Artificial Intelligence - XAI}

Based on the growing need to gain confidence in black-box models, the XAI community has proposed different methodologies, techniques and tools to explain these models. It is argued that, based on the creation of model explanation layers, a human user can create their interpretations and thus better understand how the model's decisions were generated, therefore obtaining greater confidence \cite{arrieta2020explainable,molnar2020interpretable_bookref}.
One of the most popular categories of XAI techniques currently available is the so-called post-hoc explanations. The main particularity of these post-hoc explanations is the fact that they only use training data, test data, model output data and the model itself, already properly trained to generate the explanations \cite{arrieta2020explainable}. One of the most current and necessary characteristics that an XAI technique can feature is the fact that it is applicable to computational models of independent structural natures (neural network, tree, vector of weights etc,...). This feature is called model-agnostic \cite{molnar2020interpretable_bookref}.

Among the current post-hoc XAI techniques, the following stand out: Text Explanations, Visual Explanations, Local Explanations, Explanations-by-Example, Explanations-by-Simplification and Feature Relevance Explanations. Out of these, this research highlights the Explanation-by-Example as a poorly explored technique by the XAI community. In fact, there is a smaller number of research works that present a clear proposal or tool that can be used in a replicable way for different real-world problems \cite{molnar2020interpretable_bookref,interpretabilidade_review2018,guidotti2018survey_review,molnar2020interpretable_review}.

Example-based explanation methods select specific instances of the dataset in order to explain the behavior of models or to explain the underlying data distribution \cite{molnar2020interpretable_bookref}. Explanations based on examples are mostly model-agnostic, since they make any model more interpretable. The most popular tool proposals for example-based explanations are: Counterfactual explanations  \cite{wachter2017counterfactual}, Adversarial examples \cite{biggio2018wild_adversal}, Prototypes  \cite{kim2016examples_critsism} and Influential instances \cite{koh2017understanding_influence}. Each of these proposals seeks to carry out the process of identifying relevant instances of the dataset, which directly, or even indirectly, explain and justify the model's output \cite{molnar2020interpretable_bookref}.

It should be clear that the aforementioned tools feature individual differences in terms of their ability to point to meaningful instances to explain an ML model. Therefore, they may provide different results even on the same dataset. This is directly linked to the base algorithm or function on which each tool is based, as well as the complexity of the model (dataset and algorithm) analyzed \cite{gunning2019darpa,ribeiro2021does}. Thus, it is understood that the proposed study of using the IRT to explain the ML model may generate merely different results when compared to other techniques mentioned previously, so it would be difficult to make an objective comparison. Furthermore, this research aims to apply the IRT to actually explore different details from the interpretation of the IRT estimators.

\subsection{Item Response Theory - IRT}

Traditionally, the number of correct answers is used to evaluate the performance of individuals in a test. However, this approach has limitations to assess the real ability of an individual. On the other hand, the IRT allows for evaluating the latent characteristics of an individual that cannot be directly observed, and it aims to present the relationship between the likelihood of an individual responding correctly to an item and their ability. One of the main characteristics of the IRT is that the core elements are the items and not the test as a whole, that is, an individual's performance is evaluated based on their ability to get certain items right in a test and not how many items they get right \cite{baker2001basics}.

The IRT is a set of mathematical models that seek to represent the probability of an individual correctly responding an item as a function of the item parameters and the respondent's skill, as the greater the individual's skill, the greater the chance of getting the item right. Dichotomous items are the most used, as it is only considered whether the item was answered correctly or not\cite{baker2001basics}. The IRT allows for simultaneous assessment of both the items and the respondents. In order to characterize the items, the following parameters are commonly considered by IRT models: Discrimination ($a_{i}$), which represents how much the item \textit{i} differentiates between good and bad respondents. The higher its value, the more discriminating the item; Difficulty ($b_{i}$), which represents how difficult an item is to be answered correctly and the higher its value, the more difficult the item; Guessing ($c_{i}$) represents the probability of a random hit or also the probability of a low-skill respondent hit the item.

To estimate item parameters, the response set of all individuals for all items to be evaluated is used. Respondents are evaluated based on the estimated ability ($\theta_{j}$) and the probability of a correct answer calculated as a function of an individual's ability and the parameters of item $i$. The logistic IRT model that uses the three parameters (the 3PL IRT model) calculates the probability of a correct answer $U_{ij}$ by the following equation:

\begin{equation}
    P(U_{ij} = 1|\theta_{j}) = c_{i} + (1 - c_{i})\frac{1}{1+ e^{-a_{i}(\theta_{j}-b_{i})}}
\end{equation}


Both item and individual parameters are simultaneously estimated using the response set, usually by maximizing the likehood of the model given the response data. The IRT can then be understood as a “magnifying glass” that allows for observing the individual's performance in a specific way on each item and for estimating a probable skill level in the area being evaluated.

\section{Methodology}

The IRT is generally applied for educational purposes, where the respondents are students and the items are test questions.
To analyze datasets and learning algorithms through IRT in the herein research, instances of a dataset were used, with items and classifiers being assumed as respondents. The 3PL-IRT model was used because it is the most complete and consistent to fit responses \cite{martinez2019item}.

\begin{figure}[t]
\includegraphics[width=\textwidth]{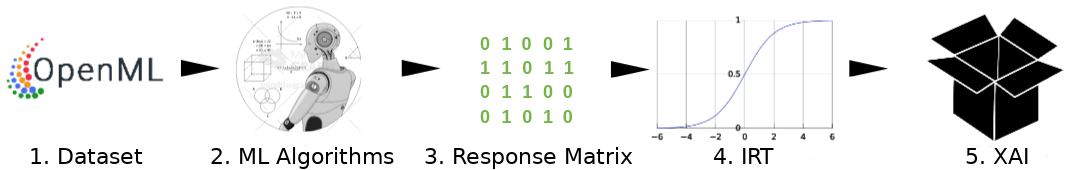}
\caption{Methodology for the application of IRT in ML and XAI.} \label{fig1}
\end{figure}

Figure \ref{fig1} illustrates the proposed methodology for applying IRT to open the box and then help explain ML models through the following steps:

\begin{enumerate}
    \item A supervised learning dataset or benchmark is chosen and divided into training and testing;
     \item Several ML models are built by using the training set and are adopted to predict the instances in the test dataset;
     \item The response from these classifiers is collected in matrix form. Each row is associated with the classifier and each column is associated with a test instance. Each matrix entry represents whether an instance was correctly classified or not by a model (a $0|1$ indicator);
     \item The response matrix is used to build the IRT model and thus to estimate the item parameters of the test instances;
     \item Finally, IRT estimators are used to open the box and assess the reliability of the model's predictions.
\end{enumerate}

\subsection{ML and IRT}

Initially, given a dataset of interest and a pool of ML algorithms, steps 1, 2 and 3 in the proposed methodology result in a matrix of responses given as input to IRT (step 4). By default, the dataset is divided in a stratified manner, being 70\% for training and 30\% for validation (step 1). In order to generate a large number of responses, three sets of classifiers were built (step 2):

\begin{enumerate}

    \item The first set is composed of 120 Random Forests models, where the number of trees gradually increases from 1 to 120;
    
    \item The second set of classifiers is: Standard Naive Bayes Gaussian, Standard Naive Bayes Bernoulli, KNN (with 2, 3, 5 and 8 Neighbors), Standard Decision Trees, Random Forest (RF) with 3 Trees, Random Forest with 5 Trees, Standard Random Forests, Standard SVM and Standard MLP. The models classified as standard mean that the standard hyperparameters of Scikit-learn \cite{pedregosa2011scikit} were used. All models are trained using 10-fold cross-validation;
    
   
   
   
   \item The third set is composed of 7 artificial classifiers advised in \cite{martinez2019item} to provide limit performance indicators of real classifiers and provide greater variability in the responses: an optimal classifier (classifies all instances correctly), a pessimal classifier (misses all classifications), a majority classifier (classifies all instances with the majority class), a minority classifier (classifies with the minority class) and three random classifiers (classifies randomly).
 
\end{enumerate}

A matrix of responses is generated based on the predictions provided by the classifiers (step 3). 
The decodIRT tool \cite{cardoso2020decoding}, which automates from step 1 to step 4, was adopted in the herein paper. By definition, the tool generates 120 MLP models as the first set of classifiers. However, as one of the objectives of this study is to explain the Random Forest model, the tool was modified to suit the research objectives. To calculate the IRT estimators (step 4), the tool depends on the item parameters calculated from the model responses, the classifiers ability and the probability of success derived from the IRT logistic model.

This research proposes that the models can be explained based on the interpretation of the IRT estimators generated in the experiments (step 5). At first, it analyzes the item parameters generated for each dataset more generically to generate a general interpretation without a specific model. Then the item parameters are analyzed considering particular characteristics of the datasets. To this end, 3D graphs and histograms are generated to understand the relationship between data and item parameters. The probability of success and the ability of the models are used to measure the confidence of the classification result by comparing it to classic ML metrics. In addition, this research intends to explore the instances from the correlation analysis between the item parameters and the vector of features that make up the data to analyze and what examples are more interesting to explain and interpret a model decision. At same time exposing the models' confidence on its decision (Is the model basically guessing?).



\subsection{Evaluated datasets}

As a case study, 4 binary datasets with different levels of complexity were chosen: Credit-g, Sonar, PC1 and Heart-Statlog. These datasets were selected from a total of 41 datasets, referring to binary classification problems extracted from OpenML \cite{vanschoren2014openml}. These datasets were selected by relying on a clustering processes that identified groups of datasets in OpenML with distinct properties. Then more varied datasets across clusters were selected. 

In the clustering process, \textit{K-means} algorithm followed by a Multiple Correspondence Analysis - MCA  \cite{ribeiro2021does} were adopted to cluster the datasets described by 15 different properties. The clustering process resulted in three main clusters, with respectively 21, 17 and 3 datasets. It is worth mentioning that this number of clusters was found from silhouette coefficient values, as recommended by the literature \cite{ref_silhouettes}. The MCA analysis also took into account the 15 different properties used in the clustering process, but with the addition of the label indicating the cluster to which each dataset belonged. Thus, as a result of the MCA, a graph was obtained with the spatial arrangement of all the analyzed datasets in relation to their 15 properties \cite{ref_mca}.

Thus, by inspecting the graph resulting from the MCA, it was possible to choose the 4 datasets mentioned previously in this topic, while taking due care to select 2 datasets from each cluster that exhibited considerable distances from one another, since datasets are therefore obtained with the most distinct properties possible. The cluster with 3 datasets was disregarded for being too small and for not showing sufficient separation from the other clusters according to the visual inspection of the MCA graph. It should be noted that the Credit-g and Heart-Statlog datasets belong to the most complex dataset cluster, while the Sonar and PC1 datasets belong to the simplest dataset cluster, as seen in \cite{ribeiro2021does}.


The Heart-Statlog is a heart disease dataset, where each instance represents a diagnosed individual whether or not you have a heart disease. The dataset has 270 instances and 13 features. The dataset also has a slight class imbalance, with 55.56\% of the instances being the majority. Sonar is a dataset of sonar signals, where each instance represents a sonar signal that has been reflected by a cylindrical rock or a metal cylinder. With 208 instances and 60 features, being 53.36\% of the majority instances. Credit-g is a dataset for credit analysis that classifies the credit risk of individuals as good or bad. It is composed of 1000 instances, with 20 features, this dataset being more unbalanced with 70\% of instances of the majority class. The PC1 dataset is a dataset and defects of the NASA Metrics Data Program, it is composed of data from the flight software for a satellite in Earth's orbit, where each instance informs whether or not the module has a defect. It has 1109 instances, with 21 features, 93\% belonging to the majority class, thus configuring a very unbalanced dataset.

\section{Results and discussion}

The evaluation of the use of IRT, regarding the Explanation-by-Example process, was split in two stages\footnote{All results can be accessed at: \url{https://github.com/LucasFerraroCardoso/IRT_XAI }}:

\begin{enumerate}
    \item The first focuses on the dataset and what explanations the item parameters can reveal about the data;
    \item The second is about the specific model generated and how the IRT estimators can act in the explanation process at the local level.
\end{enumerate}





\subsection{Datasets through the lens of IRT}

First, only the item parameters that were estimated for the test instances of the datasets will be evaluated. In IRT, discrimination and difficulty values can range from $-\infty$ to $+\infty$. Thus, in order to consider whether the items have high values of discrimination and difficulty, the established assessment value was zero (0). Thus, instances are considered very difficult and very discriminative if their respective values are greater than 0. For the guessing parameter, the limit presented by \cite{cardoso2020decoding} was used, which considers that instances with high guessing values are those with values greater than or equal to 0.2. Despite the difficulty and discrimination parameters being the most directly linked to the data, due to their characteristics, it is understood that the guessing parameter is important to consider for an indirect evaluation of the model. In view of this, the following data were computed.

As can be seen in Table \ref{tab1}, all datasets can be considered as being very discriminative because they feature a high percentage of instances with discrimination above 0. This means that the datasets can discriminate high and low skill classifiers. Therefore, models that feature a high hit rate for these datasets, indeed, can be considered skillful.
\begin{table}[!h]
  \centering
  \caption{Table with the percentage of test instances with high values of discrimination, difficulty and guessing.}~\label{tab1}
   \def\arraystretch{1.1}
  \begin{tabular}{l|c|c|c}
    \hline
    \textbf{Dataset}
     & \textbf{Discrimination}
      & \textbf{Difficulty}
    & \textbf{Gessing} \\
    \hline
    Sonar    & 87.30\%       & 4.76\%       & 14.29\%      \\
    PC1    & 93.99\%       & 2.1\%       & 3.9\%      \\
    Heart-Statlog    & 85.19\%       & 3.7\%       & 25.93\%      \\
    Credit-g    & 77.67\%       & 6\%       & 15\%      \\

    \hline
  \end{tabular}
\end{table}

The difficulty parameter can also reveal important information. In this case, all datasets have few instances with high difficulty values, this can mean that the datasets themselves are easy to classify and are not a challenge. To assess model confidence, this information can be interpreted in two ways: first, considering that a dataset represents the real world very well, with more than 90\% of the instances being considered easy, skilled models trained with that dataset have high chances of being reliable and correctly hitting new cases. However, if the dataset is not a reliable representation of the real world, this could also mean that few truly challenging cases are addressed by the dataset and thus the resulting model would only be prepared to correctly classify the easier cases. Thus, one can explain the model as being of high precision, but only for easy cases.

The guessing parameter is still difficult to assess. Regarding the application of IRT in ML, no research work was found that has deeply explored the impact of the guessing parameter; but, by using the IRT concepts, it is possible to raise some hypotheses.
In IRT, high guesswork values usually mean that there is something in the item itself that gives a ``hint'' to the low-skill respondent on what the correct answer is. In ML, this may mean that within the dataset the data may have some bias that facilitates its correct classification. This may be related to the concept of ``shortcut learning'' \cite{geirhos2020shortcut} that happens during training when the model finds a gven characteristic in the data that correlates with the correct class and then the model starts using this shortcut instead of evaluating the entire data. Thus, if the model has low skill, then its correct classification may be biased by the data and this may not be repeated in the real world as the model would not have generalized properly. An unskilled model for high-guess data would be unreliable. Future research would involve exploring this condition through the purposeful insertion of such biases and then evaluate with the IRT.


In order to deepen the explanation of the datasets by the IRT, the specific characteristics of the datasets will be considered. It is noted that the Credit-g and Heart-Statlog datasets, when compared to the Sonar and PC1 datasets, are on average less discriminative, more difficult and have a greater chance of casual accuracy. Even if by little difference, this corroborates the classification of \cite{ribeiro2021does} as being more complex. However, it is clear that the Sonar dataset, considered less complex, has the second highest percentage of difficulty and this may be related to the high dimensionality of the dataset.

Furthermore, it is understood that other metadata can also help explain a model. Although all datasets have high discrimination values, the reason these values can be different for each dataset and metadata can help reveal this difference. For the very unbalanced Credit-g and PC1 datasets, it can be seen that the percentage of very discriminative instances is very close to the percentage of the majority class, with PC1 having 93\% of instances of the majority class and 93.99\% of discrimination and Credit-g, which is composed of 70\% of the majority class and has 77.67\% of very discriminating instances.

\begin{figure*}[!htb]
\centering
\begin{subfigure}{.45\textwidth}
  \label{fig2:a}
  \raggedleft
	\includegraphics[width=1\columnwidth]{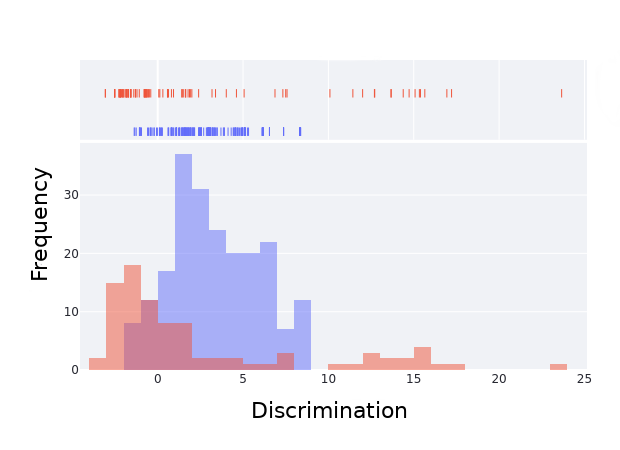} \\
	\caption{Credit-g}
\end{subfigure}%
\space \space \space 
\begin{subfigure}{.45\textwidth}
  \label{fig2:b}
  \raggedright
	\includegraphics[width=1.1\columnwidth]{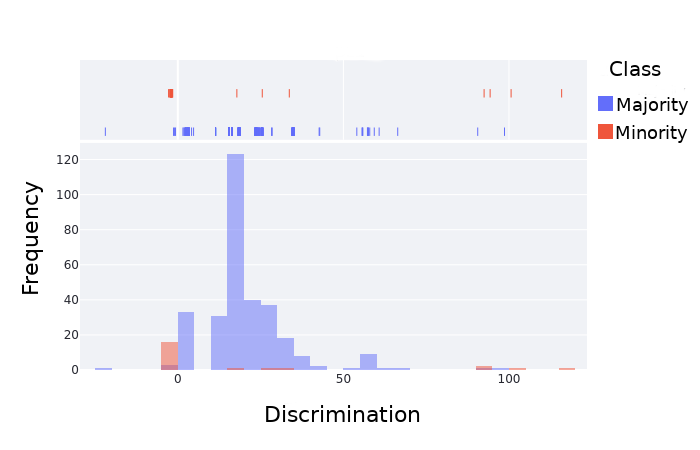} \\
	\caption{PC1}
\end{subfigure}
\caption{Discrimination Histogram separated by majority and minority class.}
\label{fig:hist}
\end{figure*}

It is common in IRT that items with high discrimination also have low difficulty, as it is understood that if a respondent makes a mistake with an item considered easy, then the ability must be low. In ML, a classifier can be considered unskilled if it cannot hit the instances referring to the majority class, as they are more recurrent in the dataset. So, it is correct to imagine that for very unbalanced datasets the percentage of very discriminative and easy instances may coincide with the majority class. As can be seen in Figure \ref{fig:hist} for the Credit-g (a) and PC1 (b) datasets, note that the histogram shows the highest number of instances with discrimination values above 0 for the majority class, while the minority class has more instances with negative discrimination. On the other hand, it is interesting that the highest discrimination values are for the minority class, this may occur due to the lower number of items and because some instances may have a strong characteristic that links them to the minority class. This can then reveal which instances of the minority class are most representative of the group and may be the most informative instances to explain the model. Besides, this information can also be useful in selecting the most suitable instances to feed oversampling techniques.
\vspace{-10pt}

\begin{figure}
\centering
\includegraphics[width=0.61\textwidth]{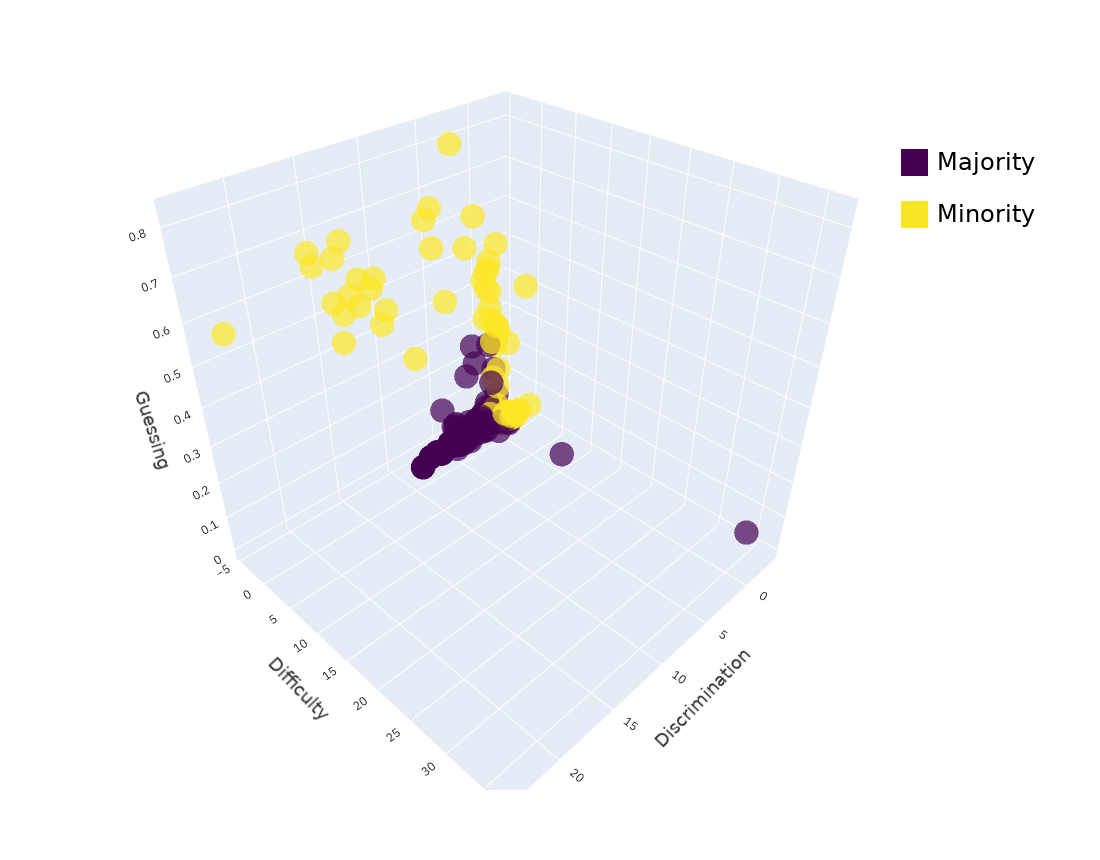}
\caption{Credit-g instances arranged over item parameters.} \label{figScatter}
\end{figure}

The Figure \ref{figScatter} shows the relationship between the majority and minority classes of the Credit-g dataset from the item parameters, it can be seen that the guessing parameter is what best distinguishes the classes, so that the minority class exhibits the more instances with high guessing values, so they are easier to hit casually. Despite assuming that this may be related to class imbalance, the same behavior is not repeated for the PC1 dataset (see supplementary material\footnotemark[7]), which is the most imbalanced one. This condition may be unique to the dataset's characteristics and would reinforce the assumption that the model may not be reliable for minority class instances if it does not have a high ability. The item parameter and model relationship will be explored in the next section.

\subsection{Random Forest through the lens of IRT}



In addition to evaluating the data in general, IRT also allows for evaluating the classifier's ability to correctly classify a specific instance. Thus, this second part of the section addresses how the IRT explains the decisions made by the evaluated model, a Random Forest with 100 trees.

Table \ref{tab2} shows the results of the test set classification of datasets by Random Forest. Due to the accuracy, a model is in place with good performance in almost all cases, only for Credit-g the model showed a lower hit rate. Considering the existing imbalance in the datasets, the Matthews Correlation Coefficient (MCC) \cite{mcc} of each model was also calculated and even for the least unbalanced datasets (Sonar and Heart-Statlog) the highest value of MCC was 0.71, indicating low correlation between classes and reinforcing the imbalance problem when considering all test instances. IRT, in turn, points to Random Forest as a skillful model, as the skill value is greater than the difficulty value in more than 90\% of the instances in all datasets. In the IRT, the respondent's skill and the item's difficulty are measured on the same scale, so that if the skill value ($\theta$) is equal to the item's difficulty, the chance of hitting must be equal to 50\% . 

\begin{table}[!h]
  \centering
  \caption{Random Forest performance for all test instances and for instances without negative discrimination.}~\label{tab2}
    \def\arraystretch{1.1}
  \begin{tabular}{l|c|c|c|c|c}
    \hline
    \textbf{Dataset}
    & \textbf{Acc Total}
    & \textbf{MCC Total}
    & \textbf{Ability $\theta$ }
    & \textbf{Acc WNG*}
    & \textbf{MCC WNG*}\\
    \hline
    Sonar   & 86\%   &  0.71    & 1.40 & 94\% & 0.88\\
PC1    & 94\%   &  0.36  & 3.76 & 99\% & 0.92\\
Heart-Statlog    & 84\%   &  0.68    & 1.20 & 97\% & 0.94\\
Credit-g      & 76\%  & 0.38   & 2.07 & 96\% & 0.88\\

    \hline
    \multicolumn{4}{l}{\small *Without Negative Discrimination.}
  \end{tabular}
\end{table}

When the difficulty limit is changed to the model's ability, it is noticed that the difficulty of the datasets has decreased considerably, reaching zero in the case of the Heart-Statlog dataset. In the specific case of this dataset, the IRT states that the generated model has a confidence of more than 50\% of success in all test instances, at least. Furthermore, the difficulty is practically zero for PC1 as well, with 0.3\% difficulty. The exception was the Sonar dataset, which kept exactly the same level of difficulty as before (4.76\%), by IRT this means that the model has less than 50\% confidence of success for 3 instances of the test set. For Credit-g the new percentage of difficult instances is 2.33\%, which means that the model has less than a 50\% hit chance for 7 instances of the test set. Such instances become very interesting to explain in what cases the model does not have a reliable prediction.

\begin{figure}
  \centering
  \includegraphics[width=.45\columnwidth]{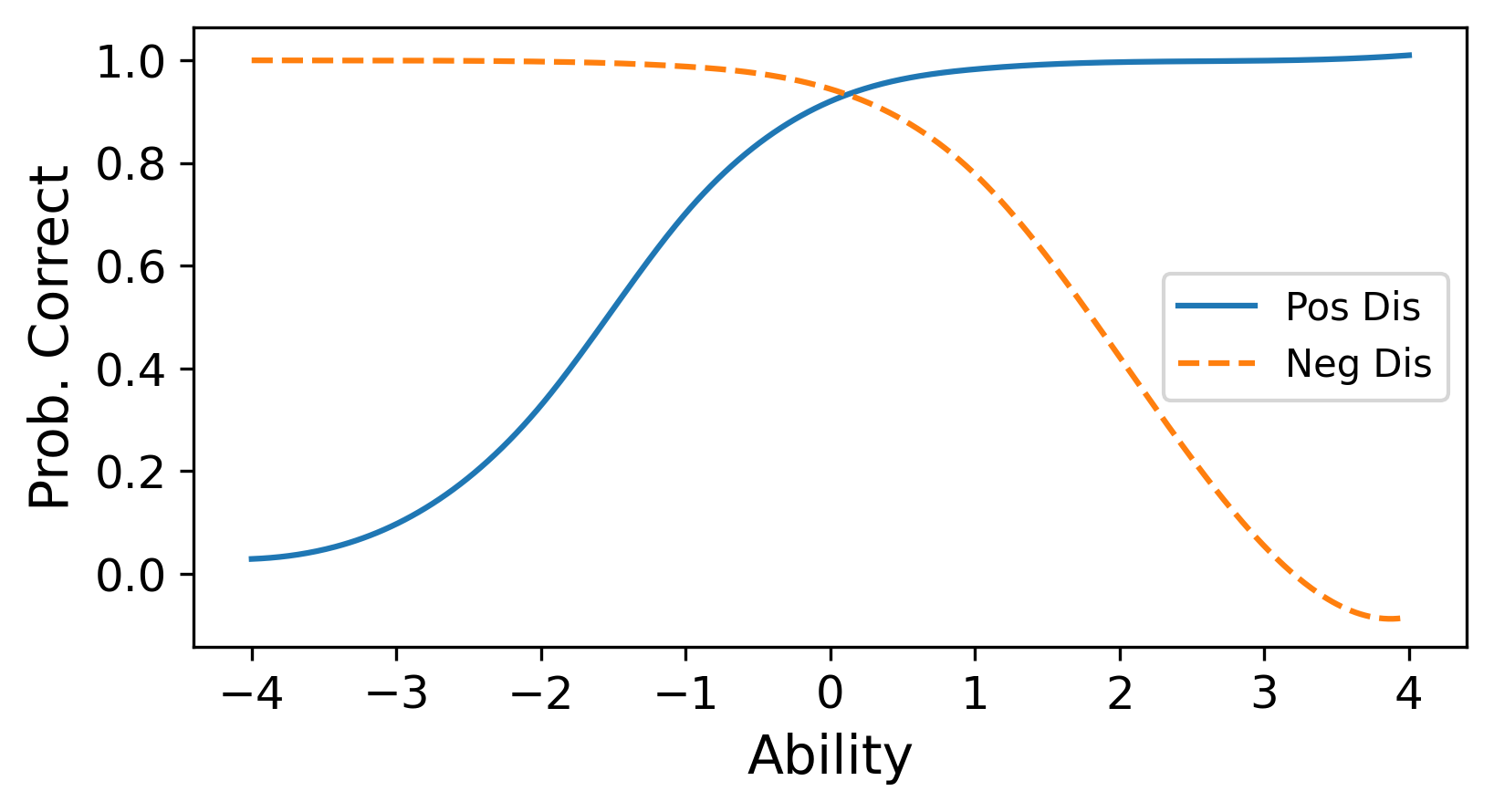}
  \caption{Comparison between items with positive and negative discrimination.}~\label{fig:icc}
\end{figure}

But if the model has such a high estimated skill and the datasets showed difficulty below the skill level of the model, then why were there still more errors and the MCC value was low? The answer to this question may also lie in the IRT discrimination parameter. In the IRT, negative discrimination values are not expected, despite being possible. The reason is that negative discrimination constitutes a situation where the less skilled respondents have the highest chance of getting it right, while the most skilled respondent has the least chance. Taking as an example two instances of Credit-g with very close values of difficulty and guessing, the probability of success of Random Forest can be completely different in both cases if the discrimination is negative. For the first instance with 1.59 discrimination the chance of success is almost 100\%, while for the instance with -1.57 discrimination the chance of success is less than 40\% (see Figure \ref{fig:icc}).

It was observed that, on average, 83.8\% of the instances that Random Forest missed in the four datasets have negative discrimination. Commonly, negative discrimination values usually mean something wrong with the item that makes it difficult to answer correctly. For the ML field, this could mean the existence of noise or an outlier in the instance. Such values may also arise when the item is different from the others, so the respondent does not have sufficient prior knowledge to answer the item correctly. For ML, this could mean that there was not enough data in training for the classifier to learn how to classify the item, as in unbalanced data correctly. Martinez et al.\cite{martinez2019item} already pointed to discrimination as a more exciting parameter than the difficulty itself.
\vspace{-10pt}

\begin{figure}
  \centering
  \includegraphics[width=0.85\columnwidth]{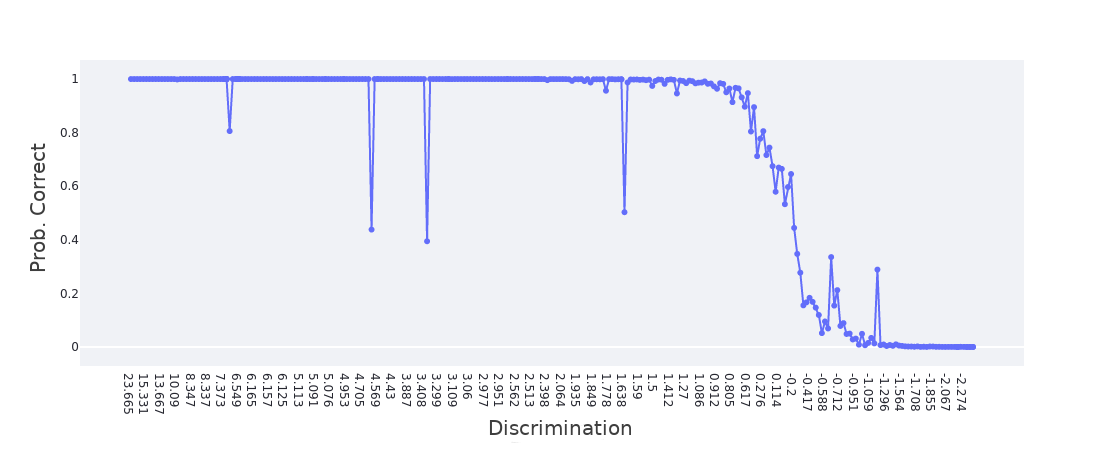}
  \caption{Probability of correct answer for each Credit-g test instance.}~\label{fig5}
\end{figure}

Figure \ref{fig5} displays the drop in the probability of success of Random Forest for the Credit-g dataset as the discrimination value of the test instances decreases; this exact configuration is repeated for the other datasets. A priori, this situation can be analyzed as follows: the model has an accuracy level above the obtained one because the instances with negative discrimination have some inconsistency and therefore can be disregarded, or the accuracy is correct, and these instances may represent types of data that the model is not able to classify correctly, as these data were not explored adequately during the training stage. In any case, this means that the model is not reliable for instances of negative discrimination.



Table \ref{tab2} reveals that the impact of negative discrimination is inherent to the type of data, since datasets of different complexity and characteristics were chosen and even so, all models showed improved performance when considering only instances with positive discrimination, for both the accuracy and the MCC value that more than doubled for the unbalanced datasets. The other model errors are usually related to the difficulty of the instance. As can be seen in Figure \ref{fig5}, even for instances with positive discrimination, in some cases the model has a low probability of success and this occurs when the difficulty of the instance exceeds the model's ability. It is interesting, then, to open the instances and study how the features relate to the item parameters.

Using the Heart-Statlog as an example because it is a context-sensitive dataset, a correlation analysis was performed between the values of the features and the item parameters of each instance. Out of all the test instances, only the ``chest'' feature has a slightly higher correlation with the difficulty parameter at 0.2485. However, when filtering the instances only for those that the model missed, it is already possible to notice greater correlations as can be seen in Table \ref{tab5}.

\begin{table}[!ht]
  \centering
  \caption{Features with correlation above 0.4 for some item parameters of the Heart-statlog dataset.}~\label{tab5}      \def\arraystretch{1.1}
  \begin{tabular}{l|c|c|c}
    \hline
    \textbf{Dataset}
    & \textbf{Discrimination}
    & \textbf{Difficulty}
    & \textbf{Guessing} \\
    \hline
    chest & 0.0285 & 0.6395     & -0.3330 \\
    fasting\_blood\_sugar & -0.4560 & 0.1320     & -0.1815 \\
    resting\_electrocardiographic       & -0.0911 & -0.3875     & 0.4592 \\
    number\_of\_major\_vessels       & -0.5391 & 0.2031     & -0.2924 \\

    \hline
  \end{tabular}
\end{table}

For misclassified instances, the ``chest'' feature has the highest correlation, with 0.6395 for difficulty, this means that this is the feature that makes it the most difficult to classify these instances. The ``chest'' feature is about the type of chest pain that a patient may have and when the specific context of the dataset is explored, it is noted that this evaluation by the IRT makes sense, as other studies have already pointed out that it is difficult to identify if chest pain is a sign of heart disease \cite{sabatine2012approach}. In addition, the ``resting\_electrocardiographic\_results'' feature was identified as the feature that most correlates with guesswork, at 0.4592. Therefore, this would be the feature that ``gives' the most clues, so that an unskilled model can get the classification of an instance right. The ``fasting\_blood\_sugar'' and ``number\_of\_major\_vessels'' features are the ones that have the highest correlation with discrimination, so these may be the instances that can best be used to discriminate poorly defined models. Trusted of the most trusted. Thus, a model that has these features as the most important to classify an instance, and the model misses, means that the model has low skill. And as expected, these same features have a high correlation with negative discrimination, as it is already known that they have a high correlation with the discrimination parameter itself. However, the ``oldpeak'' feature also presented a high correlation, at 0.4495 for negative discrimination, which indicates that this feature may be responsible for the composition of poorly formulated instances that impair the performance of the model.
When performing a percentile analysis, it was seen that 90\% of the test instances have less than half of the maximum possible value of this feature, both for the majority and minority classes, this situation indicates that any value above half the maximum value can turn out to be an outlier, thus bringing about an inconsistency in the instances and resulting in negative discrimination. This does not mean that this or the other features indicated by the IRT should be removed from the dataset, but that it is important to be aware of their values and aware of their impact on the model's confidence.



\section{Final Considerations}

This research paper presented how the IRT can be used in the Explanation-by-Example process, aiming to assist in the process of explaining a black-box model with a focus on explaining the decision made by the model and thus greater reliability for the end user. To this end, four binary datasets of different complexity were used: Heart-statlog, Credit-g, Sonar and PC1. Along with the Random Forest black-box classifier as a case study. It was observed that the IRT is able to provide new pertinent information about the classifier and data relationship, where the item parameters can be used to evaluate if the dataset concerned really encompasses all types of cases and if its own data composition can make or break a model's classification. It was also observed that the calculation of the IRT success probability can be used to measure the level of reliability that one can have on a classifier, when the model is faced with a specific instance, and thus indicating in what specific cases the model is or is not reliable, where in 83.8\% of the wrongly classified instances the IRT points out that the model is not reliable. Future research would further explore what conditions within the instances make them have a higher or lower difficulty and discrimination, in order to create conditional rules that can predict how the model will behave in view of a new instance. 


\begin{thebibliography}{8}


\bibitem{ref_mca}
Abdi, Hervé, and Dominique Valentin. ``Multiple correspondence analysis." Encyclopedia of measurement and statistics 2.4 (2007): 651-657.

\bibitem{arrieta2020explainable}
Arrieta, Alejandro Barredo, et al. ``Explainable Artificial Intelligence (XAI): Concepts, taxonomies, opportunities and challenges toward responsible AI." Information fusion 58 (2020): 82-115.

\bibitem{baker2001basics}
Baker, Frank B. ``The basics of item response theory". For full text: http://ericae.net/irt/baker., 2001.

\bibitem{biggio2018wild_adversal}
Biggio, Battista, and Fabio Roli. ``Wild patterns: Ten years after the rise of adversarial machine learning." Pattern Recognition 84 (2018): 317-331.

\bibitem{cardoso2020decoding}
Cardoso, Lucas FF, et al. ``Decoding machine learning benchmarks." Brazilian Conference on Intelligent Systems. Springer, Cham, 2020.

\bibitem{mcc}
Chicco, Davide, and Giuseppe Jurman. ``The advantages of the Matthews correlation coefficient (MCC) over F1 score and accuracy in binary classification evaluation." BMC genomics 21.1 (2020): 1-13.

\bibitem{geirhos2020shortcut}
Geirhos, Robert, et al. ``Shortcut learning in deep neural networks." Nature Machine Intelligence 2.11 (2020): 665-673.

\bibitem{interpretabilidade_review2018}
Gilpin, Leilani H., et al. ``Explaining explanations: An overview of interpretability of machine learning." 2018 IEEE 5th International Conference on data science and advanced analytics (DSAA). IEEE, 2018.

\bibitem{gohel2021explainable}
Gohel, Prashant, Priyanka Singh, and Manoranjan Mohanty. ``Explainable AI: current status and future directions." arXiv preprint arXiv:2107.07045 (2021).

\bibitem{guidotti2018survey_review}
Guidotti, Riccardo, et al. ``A survey of methods for explaining black box models." ACM computing surveys (CSUR) 51.5 (2018): 1-42.

\bibitem{gunning2019darpa}
Gunning, David, and David Aha. ``DARPA’s explainable artificial intelligence (XAI) program." AI magazine 40.2 (2019): 44-58.

\bibitem{ref_silhouettes}
Rousseeuw, Peter J. ``Silhouettes: a graphical aid to the interpretation and validation of cluster analysis." Journal of computational and applied mathematics 20 (1987): 53-65.

\bibitem{kim2016examples_critsism}
Kim, Been, Rajiv Khanna, and Oluwasanmi O. Koyejo. ``Examples are not enough, learn to criticize! criticism for interpretability." Advances in neural information processing systems 29 (2016).

\bibitem{koh2017understanding_influence}
Koh, Pang Wei, and Percy Liang. ``Understanding black-box predictions via influence functions." International conference on machine learning. PMLR, 2017.

\bibitem{review_xai_2021}
Linardatos, Pantelis, Vasilis Papastefanopoulos, and Sotiris Kotsiantis. ``Explainable ai: A review of machine learning interpretability methods." Entropy 23.1 (2020).


\bibitem{martinez2019item}
Martínez-Plumed, Fernando, et al. ``Item response theory in AI: Analysing machine learning classifiers at the instance level." Artificial intelligence 271 (2019): 18-42.

\bibitem{molnar2020interpretable_bookref}
Molnar, Christoph. Interpretable machine learning. Lulu.com, 2020.

\bibitem{molnar2020interpretable_review}
Molnar, Christoph, Giuseppe Casalicchio, and Bernd Bischl. ``Interpretable machine learning–a brief history, state-of-the-art and challenges." Joint European Conference on Machine Learning and Knowledge Discovery in Databases. Springer, Cham, 2020.

\bibitem{trust}
Naiseh, Mohammad, et al. ``Explainable recommendation: when design meets trust calibration." World Wide Web 24.5 (2021): 1857-1884.

\bibitem{regulation2018general}
Regulation, General Data Protection. ``General data protection regulation (GDPR)." Intersoft Consulting, Accessed in October 24.1 (2018).

\bibitem{pedregosa2011scikit}
Pedregosa, Fabian, et al. ``Scikit-learn: Machine learning in Python." the Journal of machine Learning research 12 (2011): 2825-2830.

\bibitem{ribeiro2021does}
Ribeiro, José, et al. ``Does Dataset Complexity Matters for Model Explainers?." 2021 IEEE International Conference on Big Data (Big Data). IEEE, 2021.

\bibitem{sabatine2012approach}
Sabatine, Marc S., and Christopher P. Cannon. ``Approach to the patient with chest pain." In: Braunwald’s Heart Disease: A Textbook of Cardiovascular Medicine. 9th ed. Philadelphia, PA: Elsevier/Saunders (2012): 1076-86.

\bibitem{vanschoren2014openml}
Vanschoren, Joaquin, et al. ``OpenML: networked science in machine learning." ACM SIGKDD Explorations Newsletter 15.2 (2014): 49-60.

\bibitem{wachter2017counterfactual}
Wachter, Sandra, Brent Mittelstadt, and Chris Russell. ``Counterfactual explanations without opening the black box: Automated decisions and the GDPR." Harv. JL \& Tech. 31 (2017): 841.

\end{thebibliography}
\end{document}